\documentclass[conference]{IEEEtran}
\IEEEoverridecommandlockouts
\usepackage{cite}
\usepackage{amsmath,amssymb,amsfonts}
\usepackage{algorithm}
\usepackage{algorithmicx}
\usepackage{graphicx}
\usepackage{textcomp}
\usepackage{xcolor}
\usepackage{bm}
\usepackage{mathrsfs}
\usepackage{subcaption}
\usepackage{array}
\usepackage[colorlinks=true, allcolors=blue]{hyperref}
\def\BibTeX{{\rm B\kern-.05em{\sc i\kern-.025em b}\kern-.08em
    T\kern-.1667em\lower.7ex\hbox{E}\kern-.125emX}}
\begin{document}

\title{Gaussian Linear Functional Manifold Method for Massive Point Cloud Data*\\
\thanks{This work is partially supported by the US National Science Foundation grant (award \# NSF-2417607) to Zhang, Kocsis, and Yang; the US National Science Foundation grant (award \# NSF-2536549) to Zhang and Fei; and USDA NIFA (award \# 2023-68012-38992) to Fei.}
}

\author{\IEEEauthorblockN{
Hong Zhao\IEEEauthorrefmark{1},
Tonglin Zhang\IEEEauthorrefmark{2},
Baijian Yang\IEEEauthorrefmark{1},
Jin Wei-Kocsis\IEEEauthorrefmark{1},
and Songlin Fei\IEEEauthorrefmark{3}
}

\IEEEauthorblockA{
\textit{\IEEEauthorrefmark{1}School of Applied and Creative Computing, Purdue University, West Lafayette, IN, USA}\\
\textit{\IEEEauthorrefmark{2}Department of Statistics, Purdue University, West Lafayette, IN, USA}\\
\textit{\IEEEauthorrefmark{3}Department of Forestry and Natural Resources, Purdue University, West Lafayette, IN, USA}\\
Emails: \{zhao1211, tlzhang, byang, kocsis0, sfei\}@purdue.edu
}
}
\maketitle
\def\eqalign#1{\null\,\vcenter{\openup\jot\ialign
              {\strut\hfil$\displaystyle{##}$&$\displaystyle{{}##}$
               \hfil\crcr#1\crcr}}\,}
               
\begin{abstract}

Reconstructing continuous terrain manifolds from massive, unstructured airborne LiDAR point clouds remains challenging in complex Wildland–Urban Interface (WUI) environments, where deep neural networks require costly point-wise annotations and non-parametric surface reconstructors often lack structural interpretability. This paper introduces the Gaussian Linear Functional Manifold (GLFM), a physics-informed statistical framework that represents continuous surface topography with deterministic linear functional bases while modeling micro-scale diffuse laser backscatter as an isotropic Gaussian process. To avoid the quadratic cost of exact constrained maximum likelihood estimation, we develop an algebraic singular value decomposition (SVD) rank-reduction algorithm that enables linear-time parameter estimation and closed-form quadric classification. Evaluated on 35.2 km² of real-world aerial LiDAR data, GLFM automatically filters ground points and extracts morphological features, achieving an ARI of 0.9933 against field-verified ground truth and outperforming four leading baselines with an out-of-core memory footprint. The framework offers a rigorous, interpretable, and scalable foundation for large-scale point cloud analytics.






\end{abstract}


\section{Introduction}
\label{sec:introduction}

Recent advancements in airborne Light Detection and Ranging (LiDAR) and autonomous aerial remote sensing have enabled the acquisition of massive, high-precision three-dimensional (3D) point clouds, providing the physical foundation for digital twin engineering, geospatial mapping, autonomous navigation, and regional ecosystem modeling \cite{shan2018topographic}. In topographically heterogeneous environments---most notably across the Wildland-Urban Interface (WUI), where residential structures intermingle with dense vegetative canopies and complex geological formations---point cloud analytics faces severe operational challenges. The primary task in parsing these massive datasets is manifold reconstruction and semantic ground filtering: decomposing unstructured, noisy, and non-uniformly distributed spatial coordinates into topologically continuous terrain manifolds and discrete above-ground morphological primitives.

Despite two decades of research, existing methods remain constrained when applied to massive, uncurated point clouds. Supervised deep-learning models require prohibitively expensive point-level annotations, while implicit surface reconstruction methods are sensitive to data sparsity and offer limited interpretability or statistical inference. Heuristic filters rely on rigid thresholds that often fail in complex terrain, and manifold-learning approaches lack explicit physical surface representations. Moreover, conventional small-ball models inadequately capture the optical characteristics of laser altimetry.

To resolve these limitations, this paper introduces the Gaussian Linear Functional Manifold (GLFM) framework, a physics-grounded, statistically principled, and computationally scalable approach to unsupervised manifold reconstruction and terrain classification. The physical foundation stems directly from the dual-scale interaction of near-infrared laser pulses (such as 1064\,nm Nd:YAG lasers) with terrestrial surfaces. At the macroscopic scale, ground surfaces and engineered structures exhibit continuous geometric regularity that can be represented by linear combinations of functional basis elements. At the microscopic scale, surface roughness, receiver noise, and atmospheric turbulence produce diffuse optical scattering that can be modeled as independent and identically distributed Gaussian perturbations orthogonal to the underlying continuous manifold.


Our method draws on the macroscopic and microscopic physical properties of LiDAR. Airborne LiDAR emits millions of laser pulses per second and measures their reflected signals to estimate distances and reconstruct terrain and objects, typically with centimeter-level accuracy. Most systems use 1064-nm infrared lasers, whose wavelength is short relative to surface roughness. Consequently, reflections are predominantly diffuse, scattering incident light in multiple directions rather than at a single specular angle.

\begin{figure}[!h]
     \centering
     \begin{subfigure}[hb]{0.24\textwidth}
         \centering
         \includegraphics[width=\textwidth]{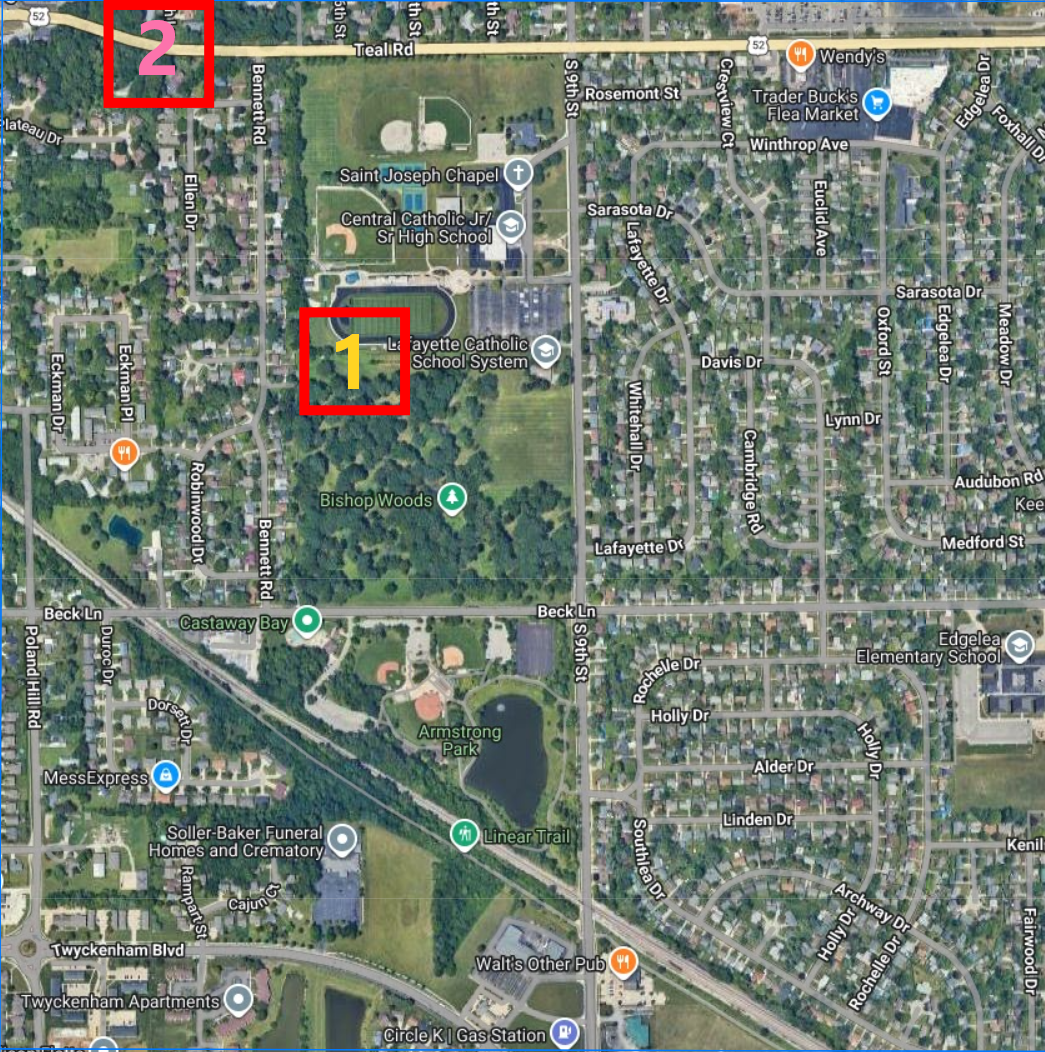} 
         \caption{Satellite Map of The Region}
     \end{subfigure}
     \begin{subfigure}[hb]{0.24\textwidth}
         \centering
         \includegraphics[width=\textwidth]{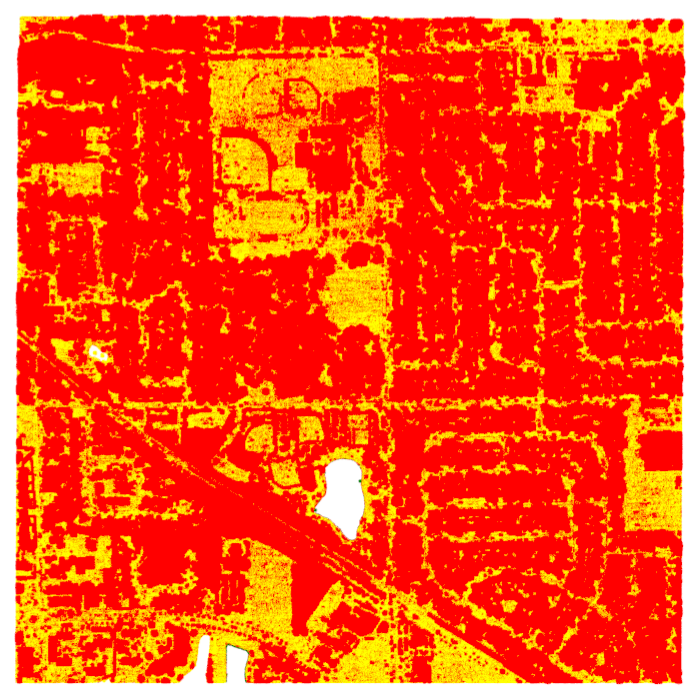} 
         \caption{Point Cloud of The Region}
     \end{subfigure}
     \begin{subfigure}[hb]{0.24\textwidth}
         \centering
         \includegraphics[width=\textwidth]{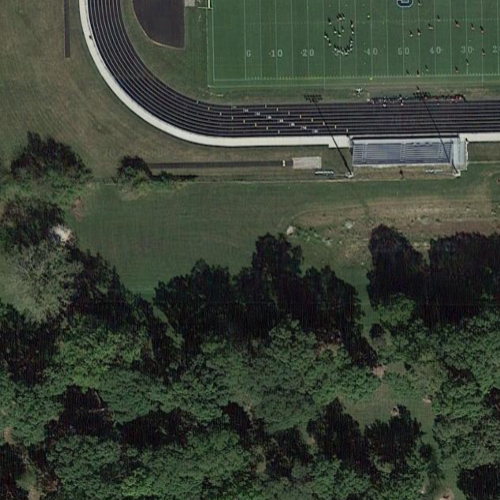} 
         \caption{Satellite Map Subarea 1}
     \end{subfigure}
     \begin{subfigure}[hb]{0.24\textwidth}
         \centering
         \includegraphics[width=\textwidth]{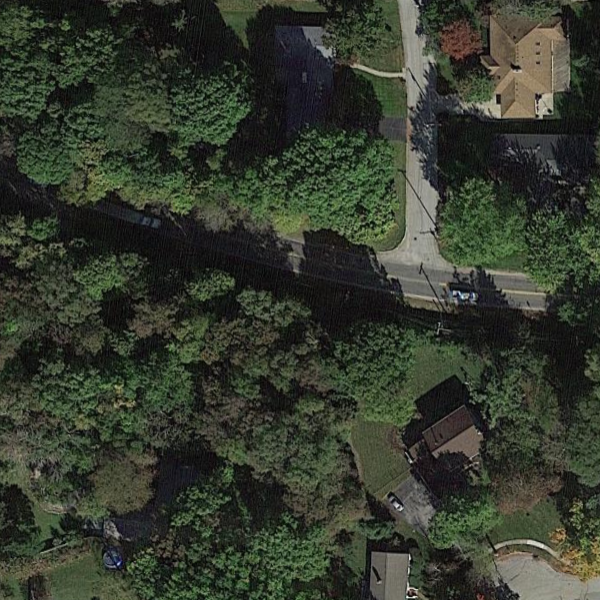} 
         \caption{Satellite Map Subarea 2}
     \end{subfigure}
    \begin{subfigure}[hb]{0.24\textwidth}
         \centering
         \includegraphics[width=\textwidth]{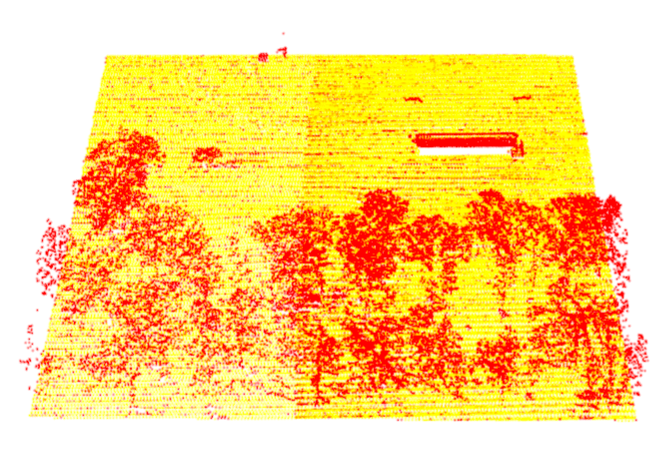} 
         \caption{Point Cloud Subarea 1}
     \end{subfigure}
     \begin{subfigure}[hb]{0.24\textwidth}
         \centering
         \includegraphics[width=\textwidth]{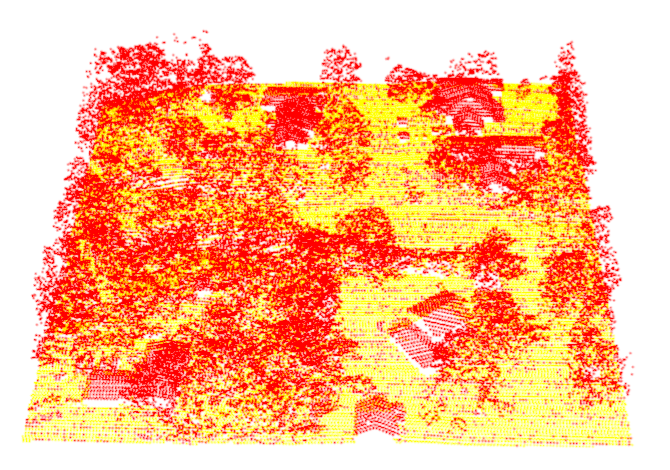} 
         \caption{Point Cloud Subarea 2}
     \end{subfigure}
    \caption{\label{fig:global_demonstration}(a) Satellite map for a $5000\times 5000({\rm ft}^2)$ region in West Lafayette, Indiana; (b) Point cloud for the region with built-in labels for {\it ground} (yellow) and {\it nonground} (red) derived by a slope-based method; (c) Satellite map of $500\times 500({\rm ft}^2)$ subarea 1; (d) Satellite map of $500\times 500({\rm ft}^2)$ subarea 2; (e) Point cloud for subarea 1; (f) Point cloud for subarea 2.}
\end{figure}



As an airborne LiDAR system traverses a wildland–urban interface (WUI), it emits laser pulses toward the surface and converts the returned signals into three-dimensional point-cloud data. The institutional repository used in this study covers an entire Midwestern U.S. state, encompassing approximately 36,421 square miles (94,326 km²), and includes both high- and low-resolution datasets. High-resolution data contain hundreds of points per square meter, whereas low-resolution data contain approximately four. The representative region shown in Fig. \ref{fig:global_demonstration}(a)–(b) contains 9,193,319 points, averaging 3.958 points per square meter. Trees, shrubs, grasses, buildings, and roads remain visually distinguishable, as further illustrated by two enlarged subareas.


At the macroscopic scale, ground and aboveground surfaces appear smooth and vary gradually across a geographic region; we represent this large-scale variation using a linear functional component. At the optical scale, however, surface roughness produces diffuse reflections with randomly varying angles, which are modeled as independent Gaussian errors across locations. The proposed method integrates these deterministic geometric and stochastic scattering components.

To capture macroscopic surface variation, we use a flexible linear functional expressed as a weighted combination of predefined basis functions, including polynomial, exponential, logarithmic, and trigonometric terms. This study focuses on polynomial bases, which can approximate a broad class of practical surface geometries. By combining the linear functional with the Gaussian error component, we formulate the model likelihood and estimate the basis coefficients using singular value decomposition (SVD)~\cite{zhang2018dimension}. The computational efficiency of SVD enables rapid model fitting.


Polynomial models also make the results easier to interpret after the point cloud is divided into ground and nonground points. For ground points, a first-order model describes a flat or sloped surface, while a second-order model captures curved terrain. The fitted coefficients can then help identify features such as ditches, depressions, and raised areas. The same approach can be applied to nonground objects.

The primary contributions of this work are summarized as follows:
\begin{itemize}
    \item \textbf{Physics-Informed Generative Manifold Formulation:} We formulate a generative modeling framework that unites macroscopic deterministic linear functional manifolds with microscopic Gaussian optical diffuse backscatter perturbations, establishing a physically accurate foundation for LiDAR surface reconstruction.
    \item \textbf{Linear-Time SVD Rank-Reduction Optimization:} We develop a computationally efficient parameter estimation algorithm that approximates the exact constrained geometric MLE via SVD rank-reduction on the functional design matrix, reducing estimation complexity to linear time $\mathcal{O}(n)$ with respect to point cardinality.
    \item \textbf{Closed-Form Quadric Spectral Morphology:} We construct an analytical framework that maps estimated quadratic tensor parameters into canonical geometric invariants, enabling unsupervised classification of geological features (flat planes, uplifts, and drainage ditches) via spectral analysis.
    \item \textbf{Likelihood Ratio Ground Filtering Pipeline:} We introduce a localized model-selection mechanism based on asymptotic log-likelihood ratio tests between linear and quadratic manifold hypotheses, achieving unsupervised separation of ground topography, structural infrastructure, and vegetative noise.
    \item \textbf{Extensive Empirical Validation:} We execute extensive validation over a regional 35.2\,km$^2$ aerial LiDAR repository, demonstrating superior classification fidelity ($\text{ARI} = 0.9933$) and computational throughput (3.528 minutes) compared to industry-standard filtering benchmarks.
\end{itemize}

The rest of the article is organized as follows. In Section~\ref{sec:related work}, we review the related work. In Section~\ref{sec:method}, we introduce our method. In Section~\ref{sec:experiment}, we present our experimental results. In Section~\ref{sec:conclusion}, we conclude.


\section{Related Work}
\label{sec:related work}

The extraction of continuous geometric structures and semantic features from unorganized three-dimensional point clouds lies at the intersection of geometric deep learning, computational graphics, remote sensing, and manifold learning.

\textbf{Supervised Deep Geometric Architectures:} Deep learning architectures for irregular point sets have achieved substantial success in 3D object classification, semantic segmentation, and part parsing. PointNet introduced permutation-invariant spatial encoding directly from point coordinates using multilayer perceptrons and symmetric pooling functions \cite{qi2017pointnet}. Despite their expressive capacity, supervised networks require large quantities of densely annotated point-level data. Producing such annotations for airborne LiDAR surveys covering extensive geographic regions is economically and logistically impractical. These models may also experience domain shift when applied to terrain that differs from their training data, motivating the development of mathematically grounded, unsupervised approaches.

\textbf{Implicit Surface Reconstruction:} In computational geometry, continuous surfaces are commonly reconstructed from unorganized point samples using implicit geometric representations. Poisson surface reconstruction, for example, estimates a continuous indicator function by solving a global Poisson equation over an estimated normal vector field \cite{kazhdan2006poisson}. Although these methods can reconstruct smooth object boundaries, they are often sensitive to nonuniform point density, sensor gaps, and multilayered vegetation. Their implicit representations also provide limited parameter interpretability and do not naturally support statistical hypothesis testing or terrain-feature classification.

\textbf{Heuristic Remote Sensing and Ground Filtering:} Ground filtering in remote sensing has traditionally relied on heuristic geometric algorithms. Slope-based filters classify steep local elevation changes as nonground structures \cite{vosselman2000slope}, while the Progressive Morphological Filter applies opening operations with progressively larger windows \cite{zhang2003progressive}. Multiscale Curvature Classification combines surface interpolation with curvature thresholds \cite{evans2007multiscale}, and Progressive TIN Densification iteratively incorporates points satisfying angular and distance constraints \cite{ni2026terrain}. Cloth Simulation Filtering instead inverts the point cloud and simulates an elastic surface draped over the terrain \cite{zhang2016easy}. Although computationally tractable, these methods depend on empirically selected thresholds that may fail in steep, uneven, or densely vegetated environments.

\textbf{Manifold Learning and Reconstruction:} A fundamental assumption of manifold learning is that high-dimensional observations lie on or near lower-dimensional manifolds embedded in Euclidean space \cite{roweis2000nonlinear,tenenbaum2000global,fefferman2016testing,hein2006manifold,narayanan2010sample}. Empirical studies have also provided evidence supporting this assumption \cite{brand2002charting}. Existing manifold-learning methods can be broadly divided into dimensionality-reduction and diffusion-based approaches.

Dimensionality-reduction methods seek low-dimensional nonlinear representations of observations from high-dimensional Euclidean spaces \cite{aamari2018stability,mcinnes2018umap,potzsche2006taylor,zhan2011robust,zhang2004principal,yao2026principal}. Representative methods include Locally Linear Embedding, Isomap, Uniform Manifold Approximation and Projection, and local tangent-space approaches \cite{roweis2000nonlinear,tenenbaum2000global,mcinnes2018umap,zhang2004principal}. Diffusion-based methods extend stochastic and generative processes from Euclidean spaces to Riemannian manifolds \cite{lin2008riemannian,de2022riemannian,jo2023generative,huang2022riemannian,lou2023scaling,braun2024riemannian,spell2023mixture}. However, these methods generally produce latent embeddings rather than explicit parametric surface equations in $\mathbb{R}^3$, limiting their interpretability in engineering applications.

Theoretical manifold reconstruction frequently relies on the concept of reach, which describes a neighborhood in which points have unique projections onto the underlying manifold \cite{calder2022boundary,niyogi2011topological}. Although useful for analyzing sampling and geometric regularity, this abstraction does not directly represent the physical reflection process of airborne LiDAR. Previous reconstruction methods also commonly rely on hyperplane approximations or estimated normal vector fields \cite{bradley2000k,kazhdan2006poisson}. Local hyperplane models improve their ability to represent nonlinear manifolds by assuming local linearity \cite{liu2023linear}, but they still lack a unified, interpretable representation of surface geometry and measurement uncertainty.

The proposed GLFM framework addresses these limitations by combining deterministic linear functional bases with Gaussian error components. This formulation provides explicit surface equations, interpretable geometric parameters, and statistical hypothesis testing while remaining computationally scalable for large LiDAR point clouds.

\section{Method}
\label{sec:method}

The physical principles of reflections of LiDAR lasers show that it is appropriate to assume that point clouds are derived by combining points on a manifold with measurement errors. We extend the physical reality and consider a more general scenario. We assume that points of a point cloud dataset reside in an Euclidean space denoted as $\mathbb{R}^r$. We express the manifold as ${\mathcal M}=\{{\bm x}^*\in\mathbb{R}^r: F({\bm x}^*)=0\}$, where $F({\bm x}^*)$ is second-order continuous. If the dimension of ${\mathcal M}$ is $r-1$, then $F(\cdot)$ is a univariate (i.e., real-valued) function. If the dimension of ${\mathcal M}$ is less than $r-1$, then $F(\cdot)$ is a multivariate (i.e., vector-valued) function. We take $r=3$ when we apply our method to 3D point cloud data collected by usual airborne LiDAR systems. If $F(\cdot)$ is univariate, then ${\mathcal M}$ is a 2D manifold. If $F(\cdot)$ is two-dimensional, then ${\mathcal M}$ is a 1D manifold. Our research problem is more general than the manifold reconstruction problem involved in 3D point clouds. 
We focus our presentation on the case when $F(\cdot)$ is univariate. We supply a way to extend it to the multivariate case at the end of this section. 

We use the normal distribution to model the errors. It means that points ${\bm x}_i=(x_{i1},\cdots,x_{ir})^\top$ in the point cloud can be expressed as
\begin{equation}
\label{eq:point cloud with Gaussian error}
{\bm x}_i={\bm x}_i^*+{\bm\epsilon}_i,{\bm\epsilon}_i\sim^{iid}{\mathcal N}({\bm 0},\sigma^2{\bf I}_d),
\end{equation}
for $i=1,\cdots,n$, where ${\bm x}_1^*,\cdots,{\bm x}_n^*\in{\mathcal M}$ are unobserved and $n$ is the number of points. We choose $F(\cdot)$ as a linear functional as
\begin{equation}
\label{eq:linear functional in the proposed GLFM model}
F({\bm x}_i^*)=\sum_{j=0}^p \beta_j\psi_j({\bm x}_i^*),
\end{equation}
where ${\bm\beta}=(\beta_0,,\dots,\beta_p)^\top\in\mathbb{R}^{p+1}$ is a parametric vector of linear coefficients to be estimated from the data and ${\bm\psi}(\cdot)=(\psi_0(\cdot),\cdots,\psi_p(\cdot))^\top$ is a vector of known basis functions. The basis functions are determined subject to interests. For instance, if we are interested in hyperplanes, then we choose $p=r$, $\psi_0({\bm x}_i^*)=1$, and $\psi_j({\bm x}_i^*)=x_{ij}^*$ for $j=1,\cdots,r$, where $x_{ij}^*$ is the $j$th component of ${\bm x}_i^*$. It specifies~\eqref{eq:linear functional in the proposed GLFM model} as
\begin{equation}
\label{eq:linear function for hyperplanes}
F_1({\bm x}_i^*)=\beta_0+\sum_{j=1}^r\beta_jx_{ij}^*.
\end{equation}
If we are interested in quadratic manifolds, then we choose $p=r(r+1)/2$, $\psi_0({\bm x}_i^*)=1$, $\psi_j({\bm x}_i^*)=x_{ij}^*$ for $j=1,\cdots,r$, and $\psi_{jk}({\bm x}_i^*)=x_{ij}^*x_{ik}^*$ for $1\le j\le k\le r$, leading to~\eqref{eq:linear functional in the proposed GLFM model} as 
\begin{equation}
\label{eq:linear function for quadratic manifold}
F_2({\bm x}_i^*)=\beta_0+\sum_{j=1}^r\beta_jx_{ij}^*+\sum_{j=1}^r\sum_{k=j}^r \beta_{jk}x_{ij}^*x_{ik}^*.
\end{equation}
The form of linear functionals defined by~\eqref{eq:linear functional in the proposed GLFM model} covers all polynomials and many well-known functions. 

In mathematics, if~\eqref{eq:linear functional in the proposed GLFM model} is satisfied, then $F$ is called a linear functional of $\psi_0,\cdots,\psi_p$, which is denoted as $F=\sum_{j=0}^p \beta_j\psi_j={\bm\beta}^\top{\bm\psi}$. The linear functional space with basis functions $\psi_0,\cdots,\psi_p$ is defined as $\mathscr{F}=\{F: F=\sum_{j=0}^p \beta_j\psi_j={\bm\beta}^\top{\bm\psi}, {\bm\beta}\in\mathbb{R}^{p+1}\}$. The space consists of univariate or multivariate functions that can be expressed as~\eqref{eq:linear functional in the proposed GLFM model}. A linear functional can be nonlinear. Any linear functional $F\in\mathscr{F}$ is a function on $\mathbb{R}^r$. It maps ${\bm x}^*\in\mathbb{R}^r$ to a real value or vector $F({\bm x}^*)$ if $\psi_0({\bm x}^*),\cdots,\psi_p({\bm x}^*)$ are univariate and multivariate functions, respectively. We use linear functionals in our method because training ${\bm\beta}$ can be accomplished by the traditional SVD approach.

The Gaussian linear functional manifold (GLFM) model given by~\eqref{eq:point cloud with Gaussian error} and~\eqref{eq:linear functional in the proposed GLFM model} has parameters ${\bm\beta}$ and $\sigma^2$. Moreover, ${\bm x}_1^*,\cdots,{\bm x}_n^*$ are also treated as parameter vectors, because they are unknown. The log-likelihood function of the model is
\begin{equation}
\label{eq:loglikelihood GLFM}
\eqalign{
\ell({\bm x}_1^*,\cdots,{\bm x}_n^*,{\bm\beta},\sigma^2)=&-{nr\over 2}\log(2\pi)-{nr\over 2}\log(\sigma^2) \cr
&-{1\over 2\sigma^2}SSQ({\bm x}_1^*,\cdots,{\bm x}_n^*,{\bm\beta}),\cr
}\end{equation}
where 
\begin{equation}
\label{eq:the SSQ term in the loglikelihood function of GLFM}
SSQ({\bm x}_1^*,\cdots,{\bm x}_n^*,{\bm\beta})=\sum_{i=1}^n \|{\bm x}_i-{\bm x}_i^*\|^2
\end{equation}
is the sum-of-squares (SSQ) of the model. The MLEs of ${\bm x}_1^*,\cdots,{\bm x}_n^*$, ${\bm\beta}$, and $\sigma^2$, denoted as $\hat{\bm x}_1^*,\cdots,\hat{\bm x}_n^*$, $\hat{\bm\beta}$, and $\hat\sigma^2$, respectively, are computed by constrained optimization 
\begin{equation}
\label{eq:constrained optimization problem for the MLE}
\hat{\bm x}_1^*,\cdots,\hat{\bm x}_n^*, \hat{\bm\beta}, \hat\sigma^2=\mathop{\arg\!\max}_{{\bm x}_1^*,\cdots,{\bm x}_n^*,{\bm\beta},\sigma^2}{\ell({\bm x}_1^*,\cdots,{\bm x}_n^*,{\bm\beta},\sigma^2)}
\end{equation}
subject to 
\begin{equation}
\label{eq:constrains for the MLE}
F(\hat{\bm x}_i^*)=\sum_{j=0}^p \hat\beta_j\psi_j(\hat{\bm x}_i^*)=0,~\forall j=1,\dots,n.
\end{equation}

Due to the constrained optimization problem, the MLEs cannot be computed using standard maximum-likelihood algorithms. Two constraints are present. The first is the {\it manifold constraint} as $\hat{\bm x}_i^*\in{\mathcal M}$ for all $i\in\{1,\dots,n\}$. The second is the {\it linear coefficient constraint}, since the choice of $F(\cdot)$ that satisfies~\eqref{eq:linear functional in the proposed GLFM model} is not unique. If ${\mathcal M}$ is satisfied by $F(\hat{\bm x}^*)=0$, then it is also satisfied by $cF(\hat{\bm x}^*)=0$ for any $c\not=0$. To ensure the uniqueness of $\hat{\bm\beta}=(\hat\beta_0,\cdots,\hat\beta_p)^\top$, we restrict $\|\hat{\bm\beta}\|=1$ with $\hat\beta_{\hat j}>0$ and $\hat j=\mathop{\arg\!\min}_{j:\hat\beta_j\not=0}j$ in our result. If it is violated, then we can modify the result to satisfy the linear coefficient constraint. We devise a Lagrange Multiplier method for the constrained optimization problem specified by~\eqref{eq:constrained optimization problem for the MLE} and~\eqref{eq:constrains for the MLE}. 

We develop our Lagrange Multiplier algorithm based on the SSQ given by~\eqref{eq:the SSQ term in the loglikelihood function of GLFM}. We propose our Lagrangian function as
\begin{equation}
\label{eq:lagrange multiplier for GLFM}
\eqalign{
{\mathcal Q}({\bm x}_1^*,\cdots,&{\bm x}_n^*,{\bm\beta},{\bm\lambda})\cr
=&\sum_{i=1}^n\left[\|{\bm x}_i-{\bm x}_i^*\|^2+\lambda_i\sum_{j=0}^p\beta_j\psi_j({\bm x}_i^*)\right],\cr
}
\end{equation}
where ${\bm\lambda}=(\lambda_1,\cdots,\lambda_n)^\top$ is the Lagrange multiplier vector that also account for the linear coefficient constraint because we can set $\lambda_i$ to be $\lambda_i/\|{\bm\beta}_j\|$ if $\beta_{\tilde j}>0$ or $-\lambda_i/\|{\bm\beta}_j\|$ if $\beta_{\tilde j}<0$ in~\eqref{eq:lagrange multiplier for GLFM}. The solution corresponding to~\eqref{eq:constrained optimization problem for the MLE} and~\eqref{eq:constrains for the MLE} is a stationary point of the Lagrange function. The gradient vector of~\eqref{eq:lagrange multiplier for GLFM} indicates that the MLEs of ${\bm x}_1^*,\cdots,{\bm x}_n^*$ and ${\bm\beta}$ satisfy
\begin{equation}
\label{eq:to lagrange multiplier for GLFM}
\eqalign{
-2({\bm x}_i-\hat{\bm x}_i^*)+\hat\lambda_i\sum_{j=0}^p \hat\beta_j\nabla\psi_j(\hat{\bm x}_i^*)=&0, \cr
\sum_{j=0}^p\hat\beta_j\psi_j(\hat{\bm x}_i^*) =&0,\cr
\sum_{i=1}^n \hat\lambda_j\psi_j(\hat{\bm x}_i^*)=&0,
}
\end{equation}
where $\nabla\psi_j(\cdot)$ represents the gradient vector of $\psi_j(\cdot)$, $\hat\lambda_i$ is the corresponding solution to $\lambda_i$ at the stationary point, the first and second equations are needed for all $i\in\{1,\cdots,n\}$, and the third equation is needed for all $j\in\{0,\dots,p\}$. Using the Hessian matrix of~\eqref{eq:lagrange multiplier for GLFM}, the computation for~\eqref{eq:to lagrange multiplier for GLFM} is obtained by a Newton-Raphson method. It provides the exact MLEs of the parameters. The SSE of the model is 
\begin{equation}
\label{eq:SSE to the GLFM}
SSE=\sum_{i=1}^n \|{\bm x}_i-\hat{\bm x}_i^*\|^2.
\end{equation}

It is inefficient to use the Lagrange Multiplier method for a massive point cloud dataset where $n$ is extremely large due to the speed and memory issues of the Newton-Raphson method. Using the traditional SVD, we devise an approximate MLE method that can efficiently compute the estimates of parameters. To apply the SVD, we construct an $n\times(p+1)$ matrix with $n\gg p+1$ as
\begin{equation}
\label{eq:matrix to apply SVD to GLFM for parameter estimation}
{\bf X}=\left(\begin{array}{cccc} \psi_0({\bm x}_1) & \psi_1({\bm x}_1) & \cdots & \psi_p({\bm x}_1) \cr  \psi_0({\bm x}_2) & \psi_1({\bm x}_2) & \cdots & \psi_p({\bm x}_2) \cr \vdots & \vdots & \ddots & \vdots \cr \psi_0({\bm x}_n) & \psi_1({\bm x}_n) & \cdots & \psi_p({\bm x}_n) \cr \end{array} \right)
\end{equation}
based on the data. We assume that ${\bf X}$ is full rank due to the randomness of the errors. We define $\hat{\bf X}$ analogously to ${\bf X}$ given by~\eqref{eq:matrix to apply SVD to GLFM for parameter estimation} by replacing ${\bm x}_1,\cdots,{\bm x}_n$ with  $\hat{\bm x}_1,\cdots,\hat{\bm x}_n$, respectively. Then, $\hat{\bf X}$ satisfies
\begin{equation}
\label{eq:restriction of modification of data matrix on manifold}
\hat{\bf X}\hat{\bm\beta}={\bf 0},
\end{equation}
implying that ${\rm rank}(\hat{\bf X})=p<{\rm rank}({\bf X})=p+1$. Thus, $\hat{\bf X}$ is rank deficient. It is a rank-$p$ matrix approximation to ${\bf X}$ under the manifold restriction specified by~\eqref{eq:linear functional in the proposed GLFM model} when ${\bm\beta}$ is estimated by $\hat{\bm\beta}$.

By the SVD, we decompose ${\bf X}$ into
\begin{equation}
\label{eq:SVD to the matrix for GLFM for parameter estimation}
{\bf X}={\bf U}{\bf D}{\bf V},
\end{equation}
where ${\bf U}=({\bm u}_0,\cdots,{\bm u}_p)$ is an $n\times(p+1)$ matrix satisfying ${\bf U}^\top{\bf U}={\bf I}_{p+1}$, ${\bf V}=({\bm v}_0,\cdots,{\bm v}_p)$ is a $(p+1)\times(p+1)$ matrix satisfying ${\bf V}^\top{\bf V}={\bf V}{\bf V}^\top={\bf I}_{p+1}$, and ${\bf D}={\rm diag}(d_0,\cdots,d_{p})$ is a $(p+1)\times(p+1)$ diagonal matrix of singular values. The singular values are assumed to be distinct and ordered such that $d_0> d_1>\cdots> d_p>0$ is satisfied. The columns of ${\bf U}{\bf D}$ are the principal components (PCs) and the columns of ${\bf V}$ are the corresponding loadings. The $j$th PC is $PC_j=d_j{\bm u}_j$. Its sample variance is $d_j^2/n$. Let 
\begin{equation}
\label{eq:PCA based on SVD to the matrix for GLFM}
{\bf X}_k={\bf X}{\bf V}_k{\bf V}_k^\top=\sum_{j=0}^k d_j{\bm u}_j{\bm v}_j^\top={\bf U}_k{\bf D}_k{\bf V}_k^\top
\end{equation}
for an integer $k\le p+1$, where ${\bf U}_k=({\bm u}_0,\cdots,{\bm u}_k)$ is $n\times(k+1)$, ${\bf D}_k={\rm diag}(d_0,\dots,d_k)$ is $(k+1)\times(k+1)$, and ${\bf V}_k=({\bm v}_0,\cdots,{\bm v}_k)$ is $(p+1)\times(k+1)$. The total variation of ${\bf X}_k$ is $\sum_{j=0}^k d_j^2$. For anj $j>k$, there is
\begin{equation}
\label{eq:remaining PCs to the PCA matrix}
{\bf X}_k{\bm v}_{j}={\bm 0}
\end{equation}
By~\cite{liu2020low}, we have
\begin{equation}
\label{eq:optimization problem for PCA}
{\bf X}_k=\mathop{\arg\min}_{\tilde{\bf X}:{\rm rank}(\tilde{\bf X})\le k+1}\|{\bf X}-\tilde{\bf X}\|^2
\end{equation}
for arbitrary $k<p+1$, where $\|{\bf A}\|=\langle{\bf A},{\bf A}\rangle={\rm tr}({\bf A}^\top{\bf A})$ stands for the Frobenius norm of matrix ${\bf A}$. If $F(\cdot)$ is univariate in~\eqref{eq:linear functional in the proposed GLFM model}, then we choose $k=p-1$ in~\eqref{eq:PCA based on SVD to the matrix for GLFM}, implying that ${\bf X}_{p-1}$ is a rank-$p$ matrix that is the closest to ${\bf X}$ without the manifold restriction.  Comparing~\eqref{eq:remaining PCs to the PCA matrix} when $k=p-1$ with~\eqref{eq:restriction of modification of data matrix on manifold}, we obtain an approximation to the MLE of ${\bm\beta}$ as
\begin{equation}
\label{eq:MLE of beta by SVD}
\hat{\bm\beta}\approx \hat{\bm\beta}_{\approx}:={\bm v}_{p}
\end{equation}
with an approximate SSE of the model as
\begin{equation}
\label{eq:approximate SSE to the GLFM}
{SSE_{\approx}}:=d_{p}^2.
\end{equation}
with the corresponding approximate MSE of the model as $\hat\sigma_{\approx}^2=SSE$.

It is suitable to use the exact MLE method given by~\eqref{eq:constrained optimization problem for the MLE} if $n$ is small or moderate. Because the size of the Hessian matrix of the Lagrange function given by~\eqref{eq:lagrange multiplier for GLFM} is $O(n^2)$, it is inappropriate to implement the exact MLE method when $n$ is large. We address this issue by the approximate MLE method specified by~\eqref{eq:MLE of beta by SVD} and~\eqref{eq:approximate SSE to the GLFM}. After $\hat{\bm\beta}_{\approx}$ is derived, we next approximately compute the value of $\hat{\bm x}_i^*$, denoted as $\hat{\bm x}_{i,\approx}^*$, for every $i\in\{1,\dots,n\}$. The algorithm can be easily derived from the first and second equations of~\eqref{eq:to lagrange multiplier for GLFM}, leading to the estimation equation for $\hat{\bm x}_{i,\approx}^*$ as
\begin{equation}
\label{eq:approximation to the ith point in lagrange multiplier for GLFM}   
\eqalign{
-2({\bm x}_i-\hat{\bm x}_{i,\approx}^*)+\hat\lambda_{i,\approx}\sum_{j=0}^p \hat\beta_{j,\approx}\nabla\psi_j(\hat{\bm x}_{i,\approx}^*)=&0, \cr
\sum_{j=0}^p\hat\beta_{j,\approx}\psi_j(\hat{\bm x}_{i,\approx}^*) =&0,\cr
}\end{equation}
where $\hat\beta_{j,\approx}$ is the $(j+1)$th component of $\hat{\bm\beta}_{j,\approx}$ and $\hat\lambda_{j,\approx}$ is the corresponding solution to $\lambda_i$. Because~\eqref{eq:approximation to the ith point in lagrange multiplier for GLFM} holds for each $i$ individually, the memory consumption of the algorithm is $O(p^2)$, which is low in the manifold reconstruction problem for point cloud data. We propose Algorithm~\ref{alg:SVD to GLFM}. 

\begin{algorithm}[tb]
\caption{\label{alg:SVD to GLFM}Approximate MLE for GLFM Using SVD when $F(\cdot)$ Given by~\eqref{eq:linear functional in the proposed GLFM model} is Univeraite}
\begin{flushleft}
\textbf{Input}: Point Cloud Data ${\mathcal D}=\{{\bm x}_i:i=1,\dots,n\}$ and Base Functions $\psi_0(\cdot),\cdots,\psi_p(\cdot)$ \\
\textbf{Output}: $\hat{\bm\beta}_{\approx}$, $SSE_{\approx}$, and $\hat{\bm x}_{i,\approx}^*$ for $i\in\{1,\cdots,n\}$
\end{flushleft}
\begin{algorithmic}[1] 
\State{Apply~\eqref{eq:SVD to the matrix for GLFM for parameter estimation} to ${\bf X}$ that is constructed by~\eqref{eq:matrix to apply SVD to GLFM for parameter estimation}}
\State{Approximately the MLE of ${\bm\beta}$ by $\hat{\bm\beta}_{\approx}$ given by~\eqref{eq:MLE of beta by SVD}, which is the last column of the loading matrix ${\bf V}$}
\State{Approximately compute SSE of the model by $SSE_{\approx}$ by~\eqref{eq:approximate SSE to the GLFM}, which is the square of the last singular value}
\State{Approximately compute the MLE of ${\bm x}_i^*$ by $\hat{\bm x}_{i,\approx}^*$ for each $i\in\{1,\cdots,n\}$ individually by an algorithm developed under~\eqref{eq:approximation to the ith point in lagrange multiplier for GLFM}}
\State{Output}
\end{algorithmic}
\end{algorithm}

We choose $r=3$ when we apply Algorithm~\ref{alg:SVD to GLFM} to 3D point cloud data. Although we develop our method for arbitrary linear functionals, we focus on the case when~\eqref{eq:linear functional in the proposed GLFM model} is specified to be first-order and second-order polynomials in this work. It is derived by setting $F(\cdot)=F_1(\cdot)$ given by~\eqref{eq:linear function for hyperplanes} and $F(\cdot)=F_2(\cdot)$ given by~\eqref{eq:linear function for quadratic manifold} under $r=3$, respectively. If $F(\cdot)=F_1(\cdot)$, then ${\mathcal M}$ is a plane, ${\bm\beta}=(\beta_0,\beta_1,\beta_2,\beta_3)^\top\in\mathbb{R}^4$, and~\eqref{eq:matrix to apply SVD to GLFM for parameter estimation} becomes 
\begin{equation}
\label{eq:first-order polynomial 3D}
 {\bf X}=\left(\begin{array}{cccc}1 &x_{11} & x_{12} & x_{13} \cr  1 &x_{21} & x_{22} & x_{23} \cr \vdots & \vdots & \vdots & \vdots \cr 1 &x_{n1} & x_{n2} & x_{n3} \cr \end{array} \right).
\end{equation}
The loading matrix ${\bf V}$ of the SVD is a $4\times 4$ orthogonal matrix. By the SVD, Algorithm~\ref{alg:SVD to GLFM} estimates ${\bm\beta}$ by $\hat{\bm\beta}_{\approx}={\bm v}_3=(v_{03},v_{13},v_{23},v_{33})^\top$ that is the last column of ${\bf V}$. The plane is estimated by $\widehat{\mathcal M}=\widehat{\mathcal M}_1$ with
\begin{equation}
\label{eq:estimates of the plane}
\eqalign{
\widehat{\mathcal M}_1=\{{\bm x}^*=&(x_1^*,x_2^*,x_3^*)^\top\in\mathbb{R}^3:\cr&  v_{03}+v_{13}x_{1}^*+v_{23}x_2^*+v_{33}x_3^*=0\}.
}\end{equation}
If $F(\cdot)=F_2(\cdot)$, then ${\mathcal M}$ is a quadratic manifold, ${\bm\beta}=(\beta_0,\beta_1,\beta_2,\beta_3,\beta_{11},\beta_{12},\beta_{13},\beta_{22}\beta_{23},\beta_{33})^\top\in\mathbb{R}^{10}$, and the corresponding ${\bf X}$ given by~\eqref{eq:matrix to apply SVD to GLFM for parameter estimation} can be constructed similarly by adding columns for the quadratic terms. Algorithm~\ref{alg:SVD to GLFM} estimates ${\bm\beta}$ by $\hat{\bm\beta}_{\approx}={\bm v}_9=(v_{09},\cdots,v_{99})^\top$ that is also the last column of the loading matrix ${\bf V}$ given by the SVD. The quadratic manifold is estimated by $\widehat{\mathcal M}=\widehat{\mathcal M}_2$ with
\begin{equation}
\label{eq:estimates of the quadratic manifold}
\eqalign{
\widehat{\mathcal M}_2=\{{\bm x}^*:&v_{09}+v_{19}x_{1}^*+v_{29}x_2^*+v_{39}x_3^*\cr
&+v_{49}x_1^{*2}+v_{59}x_{1}^*x_2^*+v_{69}x_{1}^*x_{3}^*\cr
&\hspace{12pt}+v_{79}x_{2}^{*2}+v_{89}x_{2}^*x_{3}^*+v_{99}^*x_{3}^{*2}=0\}.
}\end{equation}
Let ${\bm b}=(v_{19},v_{29},v_{39})^\top$ and
\begin{equation}
\label{eq:matrix for the standard form of a quadratic manifold}
{\bf B}=\left(\begin{array}{ccc} v_{49} &  {v_{59}\over 2} &  {v_{69}\over 2} \cr {v_{59}\over 2}  & v_{79} & {v_{89}\over 2} \cr {v_{69}\over 2} & {v_{89}\over 2} & v_{99}\end{array}  \right).
\end{equation}
If ${\bf B}$ is nonsingular, then we transform $\widehat{\mathcal M}_2$ to its standard form as
\begin{equation}
\label{eq:estimates of the quadratic manifold}
\eqalign{
\widehat{\mathcal M}_2=\{{\bm x}^*: ({\bm x}^*-{\bm a})^\top{\bf A}^{-1}({\bm x}^*-{\bm a})=1\}
},
\end{equation}
where ${\bm a}=-{\bf B}^{-1}{\bm b}/2$ and ${\bf A}=({\bm a}^\top{\bf B}{\bm a}-v_{09}){\bf B}^{-1}$. The manifold $\widehat{\mathcal M}_2$ is well-defined if at least one eigenvalue of ${\bf A}$ is positive. It is an ellipsoid if all eigenvalues of ${\bf A}$ are positive or a hyperboloid otherwise. We cannot use~\eqref{eq:estimates of the quadratic manifold} to describe the quadratic manifold if ${\bf B}$ is singular. In this case, $\widehat{\mathcal M}_2$ is a paraboloid, where the interpretation is complicated. 

If~\eqref{eq:estimates of the quadratic manifold} is suitable, then we use ${\bf A}$ and ${\bm a}$ to interpret the quadratic manifold. If the third component (i.e., the elevation value) of ${\bm a}$ is large, then $\widehat{\mathcal M}_2$ is a ditch or a depression; otherwise it is an uplift. We illustrate this issue in Section~\ref{sec:experiment}. 

We compare the local similarity of a point cloud by applying our method to neighborhoods of points. We devise a likelihood ratio statistic to fulfill the research task. In particular, we replace $SSQ$ with $SSE_{\approx}$ in~\eqref{eq:loglikelihood GLFM} and then maximize the resulting function with respect $\sigma^2$. We obtain 
\begin{equation}
\label{eq:approximate maximum of the loglikelihood function}
\eqalign{
\max\ell({\bm x}_i^*,\cdots,{\bm x}_n^*,{\bm\beta},\sigma^2)\approx&-{nr\over 2}\log{2\pi e SSE_{\approx}\over nr}.
}\end{equation}
We collected datasets ${\mathcal D}_1$ and ${\mathcal D}_2$ from neighborhoods of any two points of the point cloud, respectively. We apply~\eqref{eq:approximate maximum of the loglikelihood function} to ${\mathcal D}_1$ and ${\mathcal D}_2$ separately and jointly, respectively, leading to two versions of the implementation of~\eqref{eq:approximate maximum of the loglikelihood function}. We compute the difference between the two versions, leading to the logarithm of a likelihood ratio statistic. We conclude that the two points are similar in their neighborhoods if the logarithm value of the likelihood ratio statistic is close to zero, or dissimilar otherwise. We use this approach to separate ground and nonground points. We can also use it to identify flat and uneven subareas, respectively. We present the details of the implementation in Section~\ref{sec:experiment}.

\section{Experiment}
\label{sec:experiment}

We implement our method on real 3D point cloud data downloaded from the website of the Institute for Digital Forestry of Purdue University at https://lidar.digitalforestry.org/. The database given by the inventory of the Institute for Digital Forestry splits 3D point clouds for the entire Indiana into many datasets, each for a $5000\times 5000{\rm ft}^2$ (about $2.323{\rm km}^2$) region. A dataset in the database contains the longitude, latitude, and elevation values of points with a few associated mark variables. One of the mark variables is the classification of ground/nonground for points derived by a slope-based method~\cite{vosselman2000slope}. To analyze a larger region, multiple datasets must be used. An example is the city of West Lafayette, Indiana, which has $35.2{\rm km}^2$ area size in total. The analysis of the entire city involves $16$ datasets from the inventory. One of them is displayed in Fig.~\ref{fig:global_demonstration}(b). To emphasize the main contributions of our method, we focus our presentation on this region with details on the two subareas displayed in Figs.~\ref{fig:global_demonstration}(e) and~\ref{fig:global_demonstration}(f). 

\begin{figure}
     \centering
     \begin{subfigure}[hb]{0.24\textwidth}
         \centering
         \includegraphics[width=\textwidth]{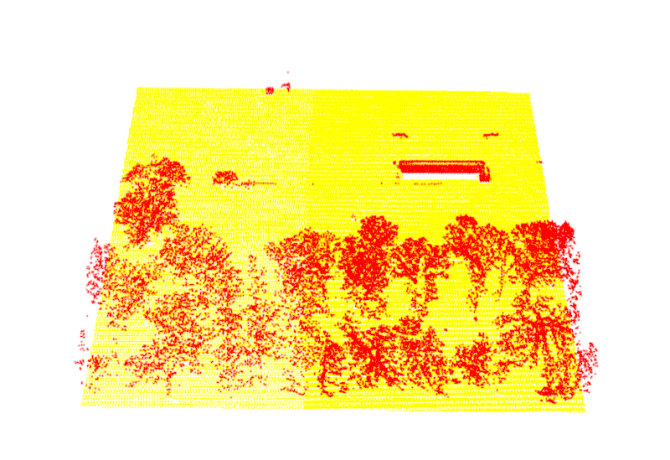} 
         \caption{Ground Truths}
     \end{subfigure}
     \begin{subfigure}[hb]{0.24\textwidth}
         \centering
         \includegraphics[width=\textwidth]{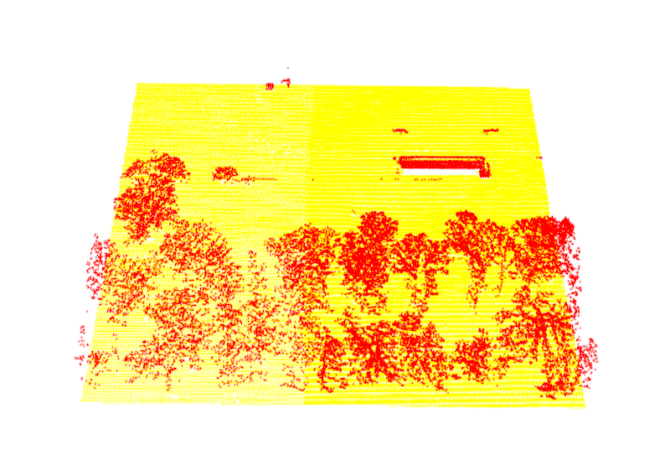} 
         \caption{Our Method}
     \end{subfigure}
     \begin{subfigure}[hb]{0.24\textwidth}
         \centering
         \includegraphics[width=\textwidth]{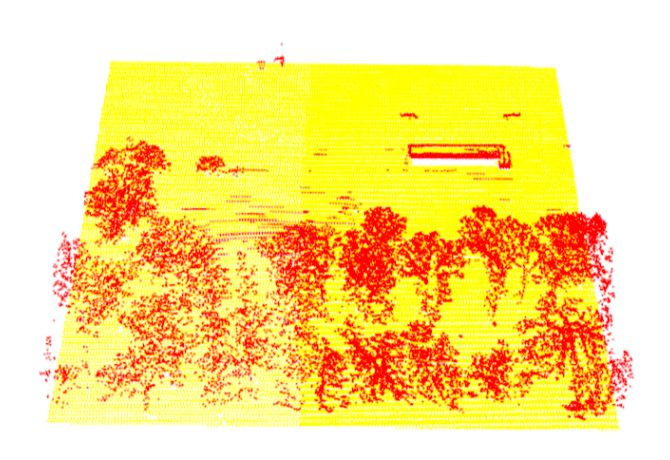} 
         \caption{CSF}
     \end{subfigure}
     \begin{subfigure}[hb]{0.24\textwidth}
         \centering
         \includegraphics[width=\textwidth]{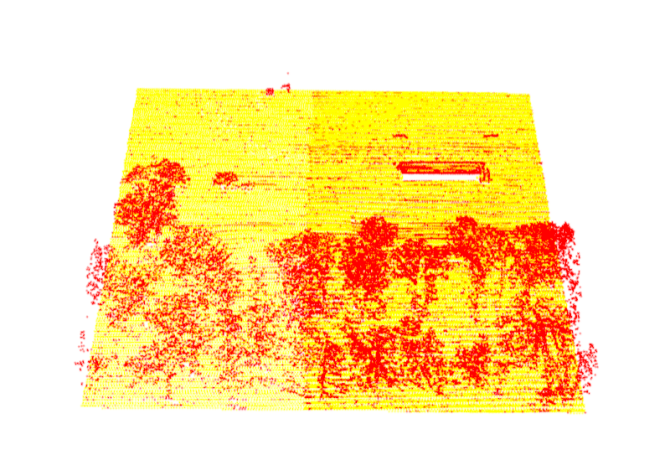} 
         \caption{PMF}
     \end{subfigure}
     \begin{subfigure}[hb]{0.24\textwidth}
         \centering
         \includegraphics[width=\textwidth]{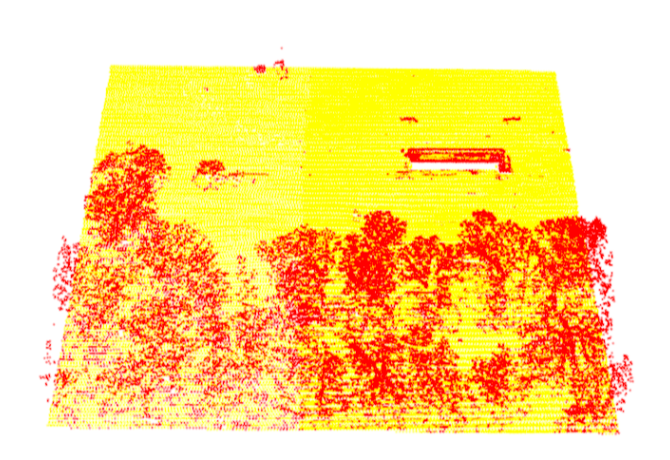} 
         \caption{MCC}
     \end{subfigure}
     \begin{subfigure}[hb]{0.24\textwidth}
         \centering
         \includegraphics[width=\textwidth]{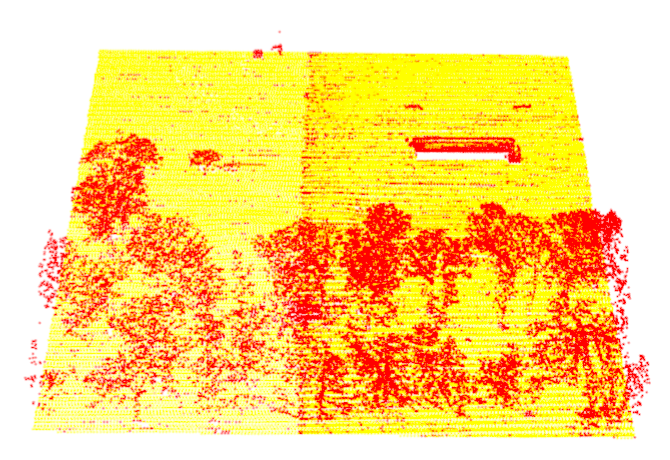} 
         \caption{PTD}
     \end{subfigure}
     \caption{\label{fig:ground_result_local_1} Ground filtering for Subarea 1 displayed in Fig.~\ref{fig:global_demonstration}(c) with the result of the slope-based method given by Fig.~\ref{fig:global_demonstration}(e). 
     }
\end{figure}

The first step to ML/AI for 3D point clouds is ground filtering. The goal is to partition data into a subset of ground points and another subset of nonground points. It assumes that ground truth is unavailable, so an unsupervised ML framework is used. The development is based on the fundamental property that nonground points are higher than ground points locally. Ground filtering is challenging even though the problem has been studied for more than twenty years. To date, well-known ground filtering methods include the slope-based~\cite{vosselman2000slope}, the Cloth Simulation Filtering (CSF)~\cite{zhang2016easy}, the Progressive Morphological Filter (PMF)~\cite{zhang2003progressive}, the Multiscale Curvature Classification (MCC)~\cite{evans2007multiscale}, and the Progressive Triangular irregular network Densification (PTD)~\cite{ni2026terrain}. The built-in labels of the database are provided by a slope-based method (e.g., Fig.~\ref{fig:global_demonstration}(b), (e), and (f)). The CSF, PMF, MCC, and PTD methods are contained in the \textsf{lidR} package of \textsf{R}. Because the difference between the built-in labels and ground truth was large, we investigated the ground filtering problem at the beginning of our work. 

To compare, we also developed our own method. In the development, we used the property that ground surfaces are smooth and their evaluation values vary slowly. It is unnecessary to use too many points to determine a ground surface. We sampled a small portion (i.e., around $1\%$) of the points from the point cloud. We then derived a neighborhood of each sampled point. We applied Algorithm~\ref{alg:SVD to GLFM} to the neighborhood with $F(\cdot)=F_1(\cdot)$ and $F(\cdot)=F_2(\cdot)$, respectively. We used a likelihood ratio test to assess whether $F(\cdot)$ was a linear or a quadratic manifold. We applied this approach to all the sampled points. It was easy to identify vegetation points because their $\hat\sigma_{\approx}^2$ values were large. We straightforwardly used the evaluation values to determine whether the remaining points belonged to buildings or the ground. We obtained subsets of the sampled points for the ground and buildings, respectively. We split the ground into two components. The first, denoted as ${\mathcal A}_1$, consisted of ground points that claimed $F(\cdot)=F_1(\cdot)$ in the likelihood ratio test. The second, denoted as ${\mathcal A}_2$, was composed of ground points that claimed $F(\cdot)=F_2(\cdot)$. We compared the estimated coefficients by the likelihood ratio test for points in ${\mathcal A}_1$. We found that we could treat them as identical in each of the $500\times 500{\rm ft}^2$ subareas. This was also correct when we made the subareas slightly larger. Because the difference between the two options could be ignored, we decided to keep using $500\times 500{\rm ft}^2$ subareas for the ground filtering problem in the city of West Lafayette in the following analysis. We then used all points in ${\mathcal A}_1$ to estimate the ground surface. We used residuals of points from the estimated ground surface to determine whether points not in the sample should be classified as ground points. We then examined points not in the sample that were close to points in ${\mathcal A}_2$. We used residuals from the estimated quadratic manifold to determine whether they should be classified as ground points. We obtained our ground filtering method. 

We applied our proposed method, the CSF, the PMF, the MCC, and the PTD ground filtering methods to the point cloud dataset displayed in Fig.~\ref{fig:global_demonstration}(c). We used ${\bm x}_i=(x_{i1},x_{i2},x_{i3})^\top$ to represent the $i$th point of the data, where $x_{i1}$ was the longitude, $x_{i2}$ was the latitude, and $x_{i3}$ was the elevation. We did not use any other variables. Based on the ground truths (Fig.~\ref{fig:ground_result_local_1}(a)) from another source, we visually compared the performance of the ground filtering methods. This shows that our method was closest to the ground truth. Many miscellaneous points appeared on the ground area when the four state-of-the-art methods were applied (Fig.~\ref{fig:ground_result_local_1}(c)-(f)). A similar phenomenon also appeared in the slope-based method (Fig.\ref{fig:global_demonstration}(e)). We compared our proposed method with the previous CSF, PMF, MCC, and PTD methods, as well as the built-in slope-based method, using the adjusted Rand Index (ARI). ARI is a popular statistical measure to quantify the similarity between two unsupervised methods. ARI values are between $0$ and $1$. A value close to $1$ indicates the greatest similarity, and vice versa. We applied ARI for paired comparison and obtained Table~\ref{table:ground_filtering_ARI_local1}. It showed that the labels assigned by the proposed method were most similar to the labels given by the ground truth. The result was almost identical to the ground truth. The performance of our method was much better than the second most similar labels assigned by the CSF method. Labels assigned by the MCC were the most dissimilar. The built-in labels given by the slope-based method adopted by the system were far from those of the ground truth, indicating that it did not work well. The previous PMF and PTD methods performed better than the previous PMF method. They still had too many errors in the ground filtering. 

\begin{table}
\caption{\label{table:ground_filtering_ARI_local1}ARI for pairwise comparison of Subarea 1 between ground truth labels, labels from the proposed ground filtering and the four previous state-of-the-art methods, as well built-in labels for Fig.~\ref{fig:ground_result_local_1}, where 1 = Ground truth, 2 = Our Method (OM), 3 = CSF, 4 = PMF, 5 = MCC, 6 = PTD, and 7 = Built-in.}
\centering
\setlength{\tabcolsep}{0.5mm}
\begin{tabular}{c|c|c|c|c|c|c|c}
\hline
\textbf{}  & \textbf{1}        & \textbf{2}        & \textbf{3} & \textbf{4} & \textbf{5} & \textbf{6} & \textbf{7} \\ \hline
\textbf{1} & 1                 & \textbf{0.995458} & 0.851821   & 0.632988   & 0.468512   & 0.709005   & 0.642193   \\ \hline
\textbf{2} & \textbf{0.995458} & 1                 & 0.850677   & 0.629885   & 0.465526   & 0.705258   & 0.638773   \\ \hline
\textbf{3} & 0.851821          & 0.850677          & 1          & 0.584711   & 0.447726   & 0.638773   & 0.577866   \\ \hline
\textbf{4} & 0.632988          & 0.629885          & 0.584711   & 1          & 0.442353   & 0.673062   & 0.618794   \\ \hline
\textbf{5} & 0.468512          & 0.465526          & 0.447726   & 0.442353   & 1          & 0.482480   & 0.448302   \\ \hline
\textbf{6} & 0.709005          & 0.705258          & 0.638773   & 0.673062   & 0.482480   & 1          & 0.696480   \\ \hline
\textbf{7} & 0.642193          & 0.638773          & 0.577866   & 0.618794   & 0.448302   & 0.696480   & 1          \\ \hline
\end{tabular}
\end{table}

\begin{figure}
\centering
    \begin{subfigure}[hb]{0.24\textwidth}
         \centering
         \includegraphics[width=\textwidth]{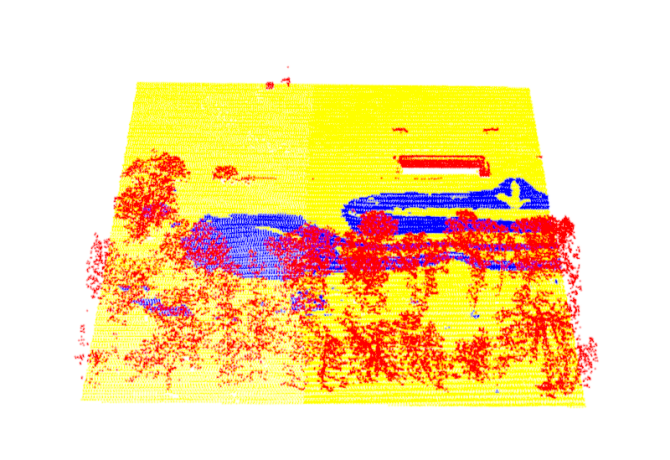} 
     \end{subfigure}
     \begin{subfigure}[hb]{0.24\textwidth}
         \centering
         \includegraphics[width=\textwidth]{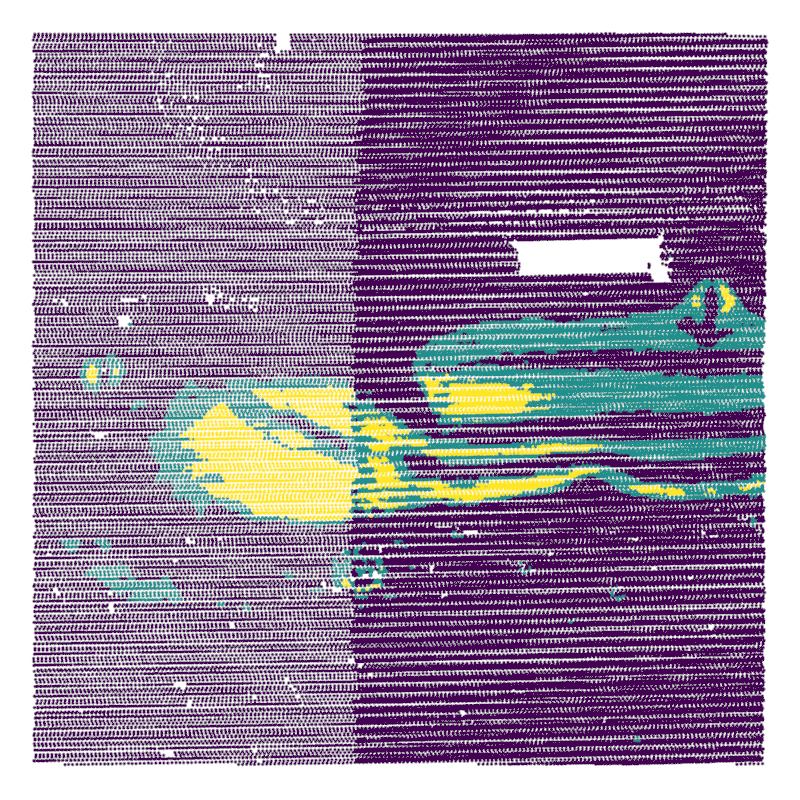} 
     \end{subfigure}
     \caption{\label{fig:Ground Shape Local 1} (a) Our method for Fig.~\ref{fig:ground_result_local_1}(b) with additional blue label for uneven ground. (b) Local ground shape identification for Fig.~\ref{fig:ground_result_local_1}(b), with dark purple indicating normally flat ground, green color indicating likely uneven ground, and yellow color indicating uneven ground for sure.}
\end{figure}

We next used our method to investigate whether ground areas were locally flat or uneven within Subarea 1. A flat ground consists of points where the manifold is locally a plane. It is determined by points where $F(\cdot)=F_1(\cdot)$ in their neighborhoods. An uneven ground consists of points where a manifold can be approximated by a second-order polynomial locally. It is determined by points with $F(\cdot)=F_2(\cdot)$ in their neighborhoods. We identified an uneven area using this property (Fig.~\ref{fig:Ground Shape Local 1}(a)). We then computed the three eigenvalues of ${\bf A}$ in the standard form of $\widehat{\mathcal M}_2$ given by~\eqref{eq:estimates of the quadratic manifold}. We connected the three eigenvalues and local curvatures of the points on the uneven ground. We classified points with large curvature values as {\it uneven for sure}, while points with small curvature values as {\it uneven likely} (Fig.\ref{fig:Ground Shape Local 1}(b)). Because evaluations of points on uneven ground were lower than those on flat ground, we found a ditch. We visually confirmed this finding. 

\begin{figure}
     \centering
     \begin{subfigure}[hb]{0.24\textwidth}
         \centering
         \includegraphics[width=\textwidth]{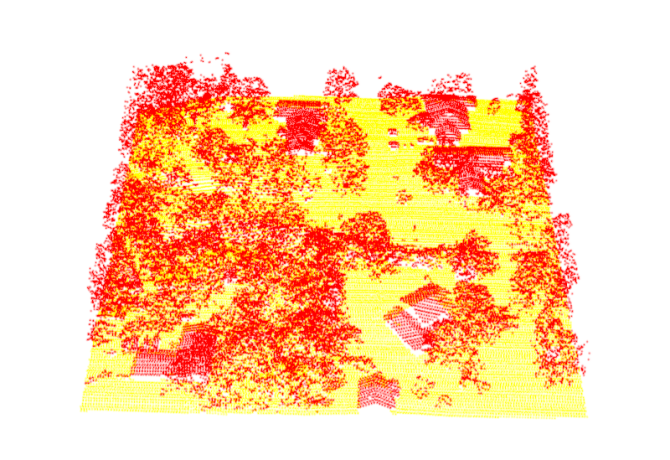} 
         \caption{Ground Truths}
     \end{subfigure}
     \begin{subfigure}[hb]{0.24\textwidth}
         \centering
         \includegraphics[width=\textwidth]{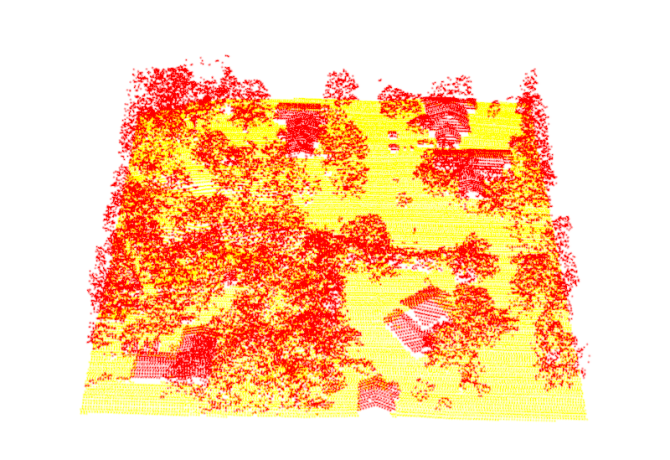} 
         \caption{Our Method}
     \end{subfigure}
     \begin{subfigure}[hb]{0.24\textwidth}
         \centering
         \includegraphics[width=\textwidth]{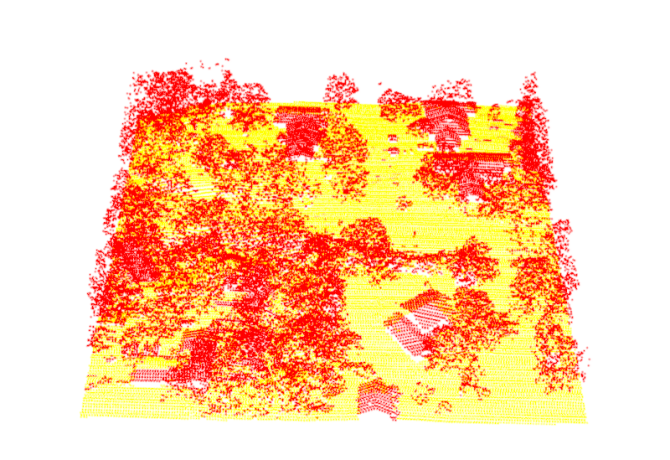} 
         \caption{CSF}
     \end{subfigure}
     \begin{subfigure}[hb]{0.24\textwidth}
         \centering
         \includegraphics[width=\textwidth]{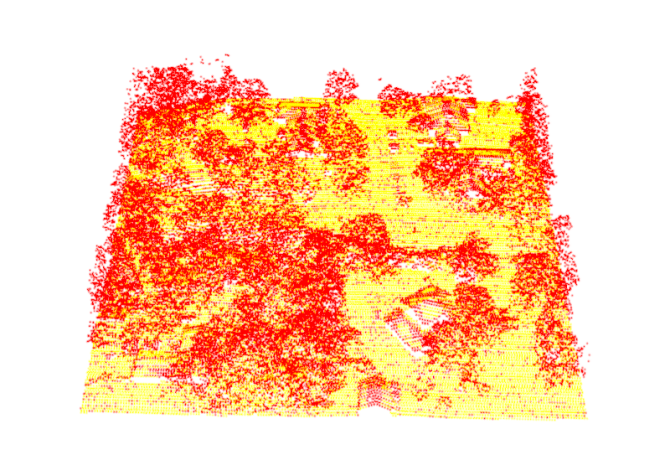} 
         \caption{PMF}
     \end{subfigure}
     \begin{subfigure}[hb]{0.24\textwidth}
         \centering
         \includegraphics[width=\textwidth]{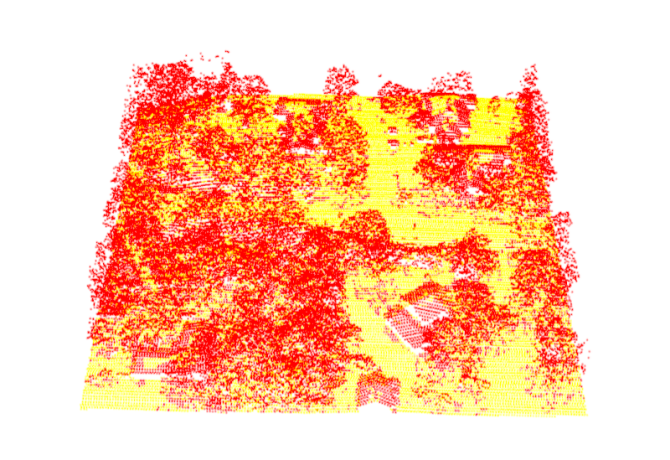} 
         \caption{MCC}
     \end{subfigure}
     \begin{subfigure}[hb]{0.24\textwidth}
         \centering
         \includegraphics[width=\textwidth]{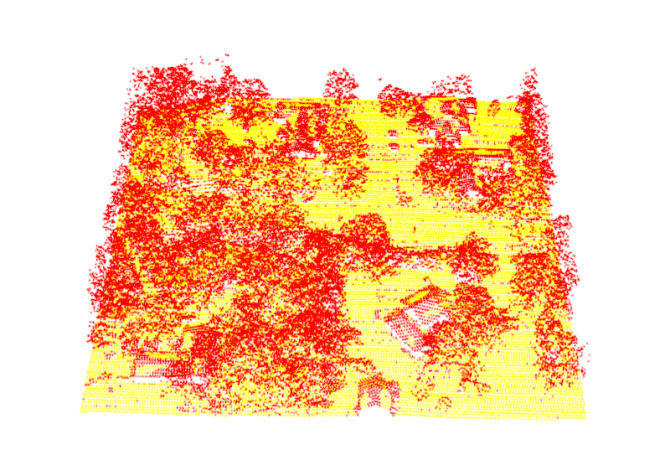} 
         \caption{PTD}
     \end{subfigure}
     \caption{\label{fig:ground_result_local_2}Ground filtering for Subarea 2 displayed in Fig.~\ref{fig:global_demonstration}(d) with the result of the sloped method given by Fig.~\ref{fig:global_demonstration}(f).}
\end{figure}

\begin{table}
\caption{\label{table:ground_filtering_ARI_local2}ARI for pairwise comparison of Subarea 2 between ground truth labels, labels from the proposed ground filtering and the four previous state-of-the-art methods, as well built-in labels for Fig.~\ref{fig:ground_result_local_2}, where 1 = Ground truth, 2 = Our Method (OM), 3 = CSF, 4 = PMF, 5 = MCC, 6 = PTD, and 7 = Built-in.}
\centering
\setlength{\tabcolsep}{0.5mm}
\begin{tabular}{c|c|c|c|c|c|c|c}
\hline
\textbf{}  & \textbf{1}        & \textbf{2}        & \textbf{3} & \textbf{4} & \textbf{5} & \textbf{6} & \textbf{7} \\ \hline
\textbf{1} & 1                 & \textbf{0.997877} & 0.943426   & 0.568766   & 0.683411   & 0.605372   & 0.489290   \\ \hline
\textbf{2} & \textbf{0.997877} & 1                 & 0.942604   & 0.567198   & 0.681667   & 0.603748   & 0.487863   \\ \hline
\textbf{3} & 0.943426          & 0.942604          & 1          & 0.550283   & 0.646264   & 0.585079   & 0.465807   \\ \hline
\textbf{4} & 0.568766          & 0.567198          & 0.550283   & 1          & 0.467762   & 0.586149   & 0.478170   \\ \hline
\textbf{5} & 0.683411          & 0.681667          & 0.646264   & 0.467762   & 1          & 0.496685   & 0.408282   \\ \hline
\textbf{6} & 0.605372          & 0.603748          & 0.585079   & 0.586149   & 0.496685   & 1          & 0.582720   \\ \hline
\textbf{7} & 0.489290          & 0.487863          & 0.465807   & 0.478170   & 0.408282   & 0.582720   & 1          \\ \hline
\end{tabular}
\end{table}

We applied our proposed method and the four state-of-the-art methods to the point cloud dataset displayed in Fig.~\ref{fig:global_demonstration}(d). Based on the ground truths given by Fig.\ref{fig:ground_result_local_2}(a), we visually compared our method with our competitors. The comparison showed that the result of our method was the closest to the ground truth. The performance of the CSF method for Subarea 2 was better than that for Subarea 1. Many miscellaneous points appeared in the results of the PMF, the MCC, the PTD, and the slope-based methods (Fig.~\ref{fig:ground_result_local_2}(d)-(f), Fig.\ref{fig:global_demonstration}(f)). The numerical comparison using the ARI indicated that the performance of our method for Subarea 2 was better than that for Subarea 1. The labels assigned by our method were almost identical to the ground truth. The ARI value of the CSF for Subarea 2 was also better than that for Subarea 1. Labels assigned by the remaining four methods were far from the ground truth. The built-in labels assigned by the slope-based method were the most dissimilar. It is more appropriate to use our method for ground filtering than our competitors in the point cloud data. 

\begin{figure}
\centering
\begin{subfigure}[hb]{0.24\textwidth}
         \centering
         \includegraphics[width=\textwidth]{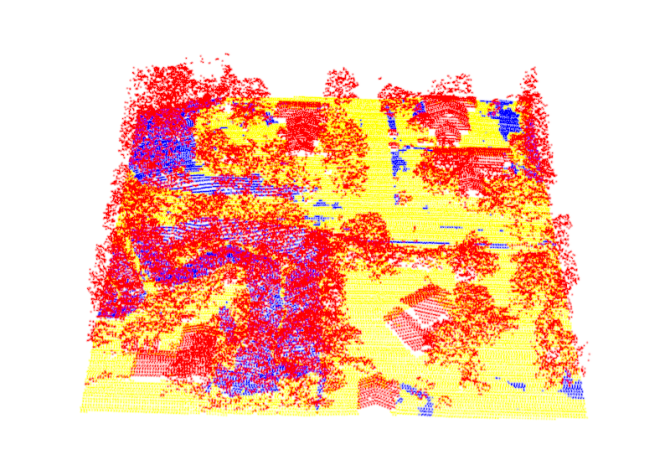} 
     \end{subfigure}
     \begin{subfigure}[hb]{0.24\textwidth}
         \centering
         \includegraphics[width=\textwidth]{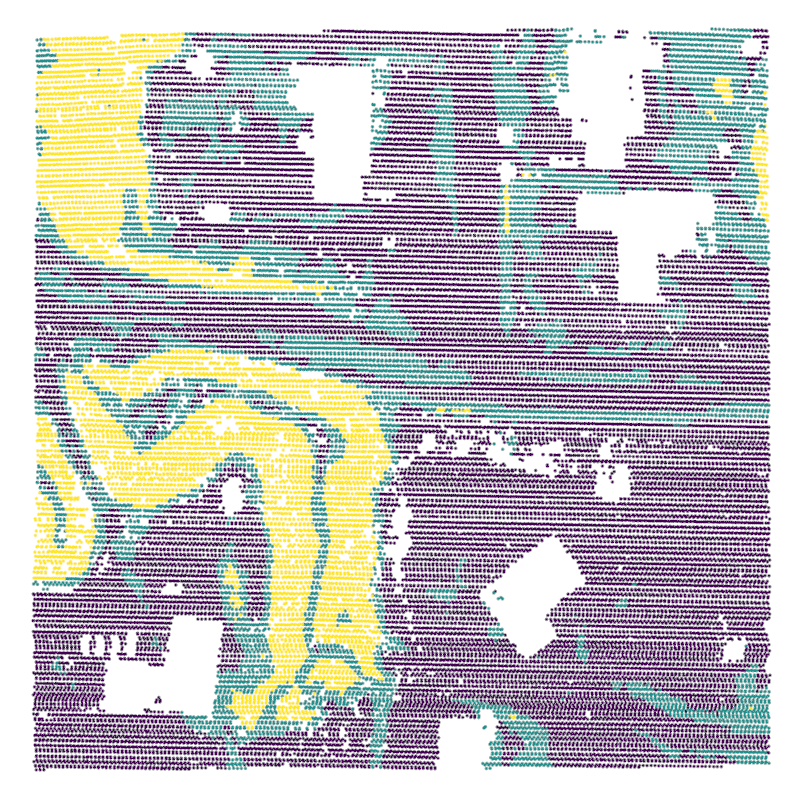} 
     \end{subfigure}
     \caption{\label{fig:Ground Shape Local 2} (a) Our method for Fig.~\ref{fig:ground_result_local_2}(b) with additional blue labels for uneven ground. (b) Ground shape identification for Fig.~\ref{fig:ground_result_local_2}(b), with dark purple indicating normally flat ground, green color indicating likely uneven ground, and yellow color indicating uneven ground for sure.}
\end{figure}

We investigated a problem about whether ground areas were locally flat or uneven in Subarea 2. We identified uneven ground by the same approach that we had used for Subarea 1. We also used the three eigenvalues of ${\bf A}$ in the standard form of $\widehat{\mathcal M}_2$ given by~\eqref{eq:estimates of the quadratic manifold} to determine uneven ground {\it for sure} and {\it likely}, respectively (Fig.\ref{fig:Ground Shape Local 2}(b)). The comparison between Figs.~\ref{fig:Ground Shape Local 2} and~\ref{fig:Ground Shape Local 1} indicating that Subarea 2 was more uneven. We confirmed this finding by visually comparing plots of point clouds for the two subareas. 

Besides Subareas 1 and 2, we studied other subareas contained in the region displayed by Figure~\ref{fig:global_demonstration}(a) with the corresponding 3D point cloud data displayed by Figure~\ref{fig:global_demonstration}(b). We consistently found that our proposed ground filtering method outperformed our competitors in general. By comparing the results of the proposed methods under $F(\cdot)=F_1(\cdot)$ and $F(\cdot)=F_2(\cdot)$, we were always able to identify flat or uneven ground, respectively. This means that an appropriate strategy for learning massive 3D point cloud data for a large geological region is to partition it into many small patches with a $500\times 500{\rm ft}^2$ area size first and then analyze the subsets for the corresponding patches individually. In this case, the ground surface can be approximated by a quadratic manifold, because higher-order terms can be ignored. We investigated this issue by using different area sizes. Our experiments showed that the results were almost identical when we slightly changed the area size for the derivation of small patches. Two obvious differences were found when we significantly changed the values. The first was that the computational burden significantly increased when we used a much smaller area size, although the corresponding results were almost identical. The second was that results changed significantly and many details were lost when we used a much larger area size, although the computation became faster. Therefore, we can approximately treat $500\times 500{\rm ft}^2$ as an optimal area size in partitioning a large geological region into small patches.

\begin{table}
\caption{\label{table:ground_filtering_ARI_global}ARI for pairwise comparison of the whole region between ground truth labels, labels from the proposed ground filtering, and the four previous state-of-the-art methods, as well built-in labels for the global, where 1 = Ground truth, 2 = Our Method, 3 = CSF, 4 = PMF, 5 = MCC, 6 = PTD, and 7 = Built-in.}
\centering
\setlength{\tabcolsep}{0.5mm}
\begin{tabular}{c|c|c|c|c|c|c|c}
\hline
\textbf{} & \textbf{1} & \textbf{2} & \textbf{3} & \textbf{4} & \textbf{5} & \textbf{6} & \textbf{7} \\ \hline
\textbf{1} & 1       & \textbf{0.993295} & 0.910675   & 0.580936   & 0.632243   & 0.672066   & 0.646942   \\ \hline
\textbf{2} & \textbf{0.993295} & 1                 & 0.912301   & 0.577985   & 0.629054   & 0.668266   & 0.643273   \\ \hline
\textbf{3} & 0.910675          & 0.912301          & 1          & 0.572880   & 0.613268   & 0.654479   & 0.587211   \\ \hline
\textbf{4} & 0.580936          & 0.577985          & 0.572880   & 1          & 0.453693   & 0.602306   & 0.531098   \\ \hline
\textbf{5} & 0.632243          & 0.629054          & 0.613268   & 0.453693   & 1          & 0.542234   & 0.472216   \\ \hline
\textbf{6} & 0.672066          & 0.668266          & 0.654479   & 0.602306   & 0.542234   & 1          & 0.602588   \\ \hline
\textbf{7} & 0.646942          & 0.643273          & 0.587211   & 0.531098   & 0.472216   & 0.602588   & 1          \\ \hline
\end{tabular}
\end{table}

We chose $500\times 500{\rm ft}^2$ as the criterion for area size for partitioning the geological region displayed by Figure~\ref{fig:global_demonstration}(a) into small patches. After they were derived, we applied our proposed method and the four state-of-the-art ground filtering methods to the small patches individually. The results of individual patches were combined. We obtained subsets of ground and nonground points, respectively, for the entire region (results not shown). We visually compared the two subsets. We found that the result of our method was almost identical to the ground truth. It was much better than the second best that was the CSF method. Many miscellaneous points were found in the results given by the remaining methods. Using their ARI values, we numerically compared the performance of our method with our competitors (Table~\ref{table:ground_filtering_ARI_global}). The ARI value of our method was almost equal to $1$, indicating that labels assigned by our method were almost identical to the ground truth. The ARI value of the CSF method for the entire region was higher than that for Subarea 1 but lower than that for Subarea 2. The ARI values of the remaining methods were low, with the lowest being the PMF method. We then studied variations of ARI among the small patches. We found that the ARI values of our method were consistently high and close to $1$ in all the patches, indicating that it could be treated as the ground truth. The ARI values given by the CSF method were mostly around $0.9$. The ARI values of the remaining methods were mostly around $0.6$, indicating that they misclassified many ground and nonground points. Because of their low ARI values, it is inappropriate to use the built-in labels to separate ground and nonground points. Ground filtering is still necessary in the analysis of the 3D point cloud data provided by the Institute for Digital Forestry at Purdue University. 

\begin{figure}
\centering
\includegraphics[width=0.52\textwidth]{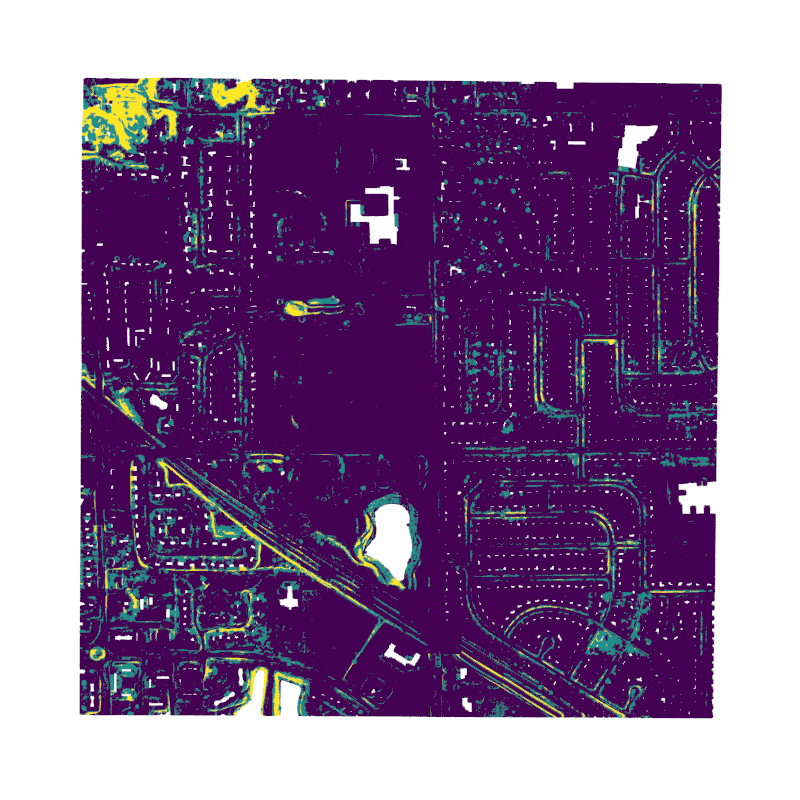}
    \caption{\label{fig:Global Ground Shape} Global ground shape identification, with dark purple indicating normally flat ground, green indicating likely uneven ground, and yellow indicating uneven ground for sure.}
\end{figure}

We applied the same method for deriving Figs.~\ref{fig:Ground Shape Local 1} and~\ref{fig:Ground Shape Local 2} to the subset of ground points given by our ground filtering method for the entire region displayed by~\ref{fig:global_demonstration}(a). We used the method to identify whether ground areas were locally flat or uneven. We obtained Figure~\ref{fig:Global Ground Shape}. We found that most of the ground areas were normally flat. A large portion of the northwestern corner of the ground was uneven. There was a spot of geological depression located at one-third of the left and one-third of the top position. We also identified a long northwest-southeast-running ditch and several downhills.  

We compared the computational efficiency of our proposed method with our competitors in the ground filtering stage. The implementation of our proposed method took $3.528$ minutes. The implementation of the CSF method took $12.799$ minutes. The implementations of the PMF, MCC, and PTD methods took $1.909$, $9.350$, and $2.739$ minutes, respectively. Note that the difference between the PMF, MCC, and PTD methods and the ground truth was large, and our method is computationally faster than the CSF method. We conclude that it is more appropriate to use our proposed method in the ground filtering stage because it is precise and computationally efficient. 

We did not compare computational efficiencies between distinct methods in the even/uneven ground detection stage because our proposed method is the only one that can fulfill the research task. We only investigated this issue for our method. The study showed that the computation for this stage was about $10$ minutes. It was slightly longer than the time taken for the ground filtering stage. 


In summary, because built-in labels are imprecise, more accurate filtering is needed to separate ground and nonground points. The proposed method performs this task while distinguishing flat from uneven terrain, demonstrating the value of manifold reconstruction for 3D point clouds. Partitioning large regions into appropriately sized patches also enables efficient processing of datasets too large to fit into system memory.





\section{Conclusion and Future Work}
\label{sec:conclusion}

This study introduced the Gaussian Linear Functional Manifold (GLFM), an interpretable and scalable framework for unsupervised manifold reconstruction, terrain classification, and ground filtering in large 3D point clouds. GLFM combines deterministic functional bases for surface geometry with Gaussian distributions representing diffuse optical scattering. To avoid costly nonlinear maximum likelihood estimation, the framework uses an SVD surrogate that supports efficient coefficient estimation and residual evaluation in linear time. It requires no manual annotations, iterative gradient descent, or extensive hyperparameter tuning.

Experiments on large airborne LiDAR datasets demonstrated that GLFM provides accurate and efficient ground filtering while distinguishing geometric features such as planar slopes, drainage ditches, depressions, and uplifts. Its explicit surface representations may also support applications in forestry, autonomous navigation, infrastructure monitoring, digital twins, and hydrological modeling.

Future work will extend the functional basis beyond low-order polynomials to model more complex geometries. Robust estimation and anisotropic covariance models will be investigated to address vegetation reflections, atmospheric interference, and nonuniform measurement errors. Additional research will examine continuity across neighboring surface patches, GPU-based parallelization, and the integration of LiDAR waveforms and hyperspectral data for large-scale geometric reconstruction and semantic analysis.

\bibliographystyle{refs/IEEEtran}
\bibliography{refs/IEEEabrv,refs/mybib}

 \newcommand{\noop}[1]{}
\begin{thebibliography}{10}
\providecommand{\url}[1]{#1}
\csname url@samestyle\endcsname
\providecommand{\newblock}{\relax}
\providecommand{\bibinfo}[2]{#2}
\providecommand{\BIBentrySTDinterwordspacing}{\spaceskip=0pt\relax}
\providecommand{\BIBentryALTinterwordstretchfactor}{4}
\providecommand{\BIBentryALTinterwordspacing}{\spaceskip=\fontdimen2\font plus
\BIBentryALTinterwordstretchfactor\fontdimen3\font minus \fontdimen4\font\relax}
\providecommand{\BIBforeignlanguage}[2]{{%
\expandafter\ifx\csname l@#1\endcsname\relax
\typeout{** WARNING: IEEEtran.bst: No hyphenation pattern has been}%
\typeout{** loaded for the language `#1'. Using the pattern for}%
\typeout{** the default language instead.}%
\else
\language=\csname l@#1\endcsname
\fi
#2}}
\providecommand{\BIBdecl}{\relax}
\BIBdecl

\bibitem{shan2018topographic}
J.~Shan and C.~K. Toth, \emph{Topographic laser ranging and scanning: principles and processing}.\hskip 1em plus 0.5em minus 0.4em\relax CRC press, 2018.

\bibitem{zhang2018dimension}
T.~Zhang and B.~Yang, ``Dimension reduction for big data,'' \emph{Statistics and Its Interface}, vol.~11, no.~2, pp. 295--306, 2018.

\bibitem{qi2017pointnet}
C.~R. Qi, H.~Su, K.~Mo, and L.~J. Guibas, ``Pointnet: Deep learning on point sets for 3d classification and segmentation,'' in \emph{Proceedings of the IEEE conference on computer vision and pattern recognition}, 2017, pp. 652--660.

\bibitem{kazhdan2006poisson}
M.~Kazhdan, M.~Bolitho, H.~Hoppe \emph{et~al.}, ``Poisson surface reconstruction,'' in \emph{Proceedings of the fourth Eurographics symposium on Geometry processing}, vol.~7, no.~4, 2006.

\bibitem{vosselman2000slope}
G.~Vosselman, ``Slope based filtering of laser altimetry data,'' \emph{International archives of photogrammetry and remote sensing}, vol.~33, no. B3/2; PART 3, pp. 935--942, 2000.

\bibitem{zhang2003progressive}
K.~Zhang, S.-C. Chen, D.~Whitman, M.-L. Shyu, J.~Yan, and C.~Zhang, ``A progressive morphological filter for removing nonground measurements from airborne lidar data,'' \emph{IEEE transactions on geoscience and remote sensing}, vol.~41, no.~4, pp. 872--882, 2003.

\bibitem{evans2007multiscale}
J.~S. Evans and A.~T. Hudak, ``A multiscale curvature algorithm for classifying discrete return lidar in forested environments,'' \emph{IEEE Transactions on Geoscience and Remote Sensing}, vol.~45, no.~4, pp. 1029--1038, 2007.

\bibitem{ni2026terrain}
W.~Ni, H.~Zhang, J.~Xu, Y.~Zhao, N.~Zheng, and W.~Shi, ``A terrain-feature-aware multi-scale ptd for airborne lidar ground filtering,'' \emph{IEEE Journal of Selected Topics in Applied Earth Observations and Remote Sensing}, 2026.

\bibitem{zhang2016easy}
W.~Zhang, J.~Qi, P.~Wan, H.~Wang, D.~Xie, X.~Wang, and G.~Yan, ``An easy-to-use airborne lidar data filtering method based on cloth simulation,'' \emph{Remote sensing}, vol.~8, no.~6, p. 501, 2016.

\bibitem{roweis2000nonlinear}
S.~T. Roweis and L.~K. Saul, ``Nonlinear dimensionality reduction by locally linear embedding,'' \emph{science}, vol. 290, no. 5500, pp. 2323--2326, 2000.

\bibitem{tenenbaum2000global}
J.~B. Tenenbaum, V.~d. Silva, and J.~C. Langford, ``A global geometric framework for nonlinear dimensionality reduction,'' \emph{science}, vol. 290, no. 5500, pp. 2319--2323, 2000.

\bibitem{fefferman2016testing}
C.~Fefferman, S.~Mitter, and H.~Narayanan, ``Testing the manifold hypothesis,'' \emph{Journal of the American Mathematical Society}, vol.~29, no.~4, pp. 983--1049, 2016.

\bibitem{hein2006manifold}
M.~Hein and M.~Maier, ``Manifold denoising,'' \emph{Advances in neural information processing systems}, vol.~19, 2006.

\bibitem{narayanan2010sample}
H.~Narayanan and S.~Mitter, ``Sample complexity of testing the manifold hypothesis,'' \emph{Advances in neural information processing systems}, vol.~23, 2010.

\bibitem{brand2002charting}
M.~Brand, ``Charting a manifold,'' \emph{Advances in neural information processing systems}, vol.~15, 2002.

\bibitem{aamari2018stability}
E.~Aamari and C.~Levrard, ``Stability and minimax optimality of tangential delaunay complexes for manifold reconstruction,'' \emph{Discrete \& Computational Geometry}, vol.~59, no.~4, pp. 923--971, 2018.

\bibitem{mcinnes2018umap}
L.~McInnes, J.~Healy, and J.~Melville, ``Umap: Uniform manifold approximation and projection for dimension reduction,'' \emph{arXiv preprint arXiv:1802.03426}, 2018.

\bibitem{potzsche2006taylor}
C.~P{\"o}tzsche and M.~Rasmussen, ``Taylor approximation of integral manifolds,'' \emph{Journal of Dynamics and Differential Equations}, vol.~18, no.~2, pp. 427--460, 2006.

\bibitem{zhan2011robust}
Y.~Zhan and J.~Yin, ``Robust local tangent space alignment via iterative weighted pca,'' \emph{Neurocomputing}, vol.~74, no.~11, pp. 1985--1993, 2011.

\bibitem{zhang2004principal}
Z.~Zhang and H.~Zha, ``Principal manifolds and nonlinear dimensionality reduction via tangent space alignment,'' \emph{SIAM journal on scientific computing}, vol.~26, no.~1, pp. 313--338, 2004.

\bibitem{yao2026principal}
Z.~Yao, B.~Eltzner, and T.~Pham, ``Principal sub-manifolds,'' \emph{Statistica Sinica}, vol.~36, pp. 1069--1089, 2026.

\bibitem{lin2008riemannian}
T.~Lin and H.~Zha, ``Riemannian manifold learning,'' \emph{IEEE transactions on pattern analysis and machine intelligence}, vol.~30, no.~5, pp. 796--809, 2008.

\bibitem{de2022riemannian}
V.~De~Bortoli, E.~Mathieu, M.~Hutchinson, J.~Thornton, Y.~W. Teh, and A.~Doucet, ``Riemannian score-based generative modelling,'' \emph{Advances in neural information processing systems}, vol.~35, pp. 2406--2422, 2022.

\bibitem{jo2023generative}
J.~Jo and S.~J. Hwang, ``Generative modeling on manifolds through mixture of riemannian diffusion processes,'' \emph{arXiv preprint arXiv:2310.07216}, 2023.

\bibitem{huang2022riemannian}
C.-W. Huang, M.~Aghajohari, J.~Bose, P.~Panangaden, and A.~Courville, ``Riemannian diffusion models,'' \emph{Advances in Neural Information Processing Systems}, vol.~35, pp. 2750--2761, 2022.

\bibitem{lou2023scaling}
A.~Lou, M.~Xu, A.~Farris, and S.~Ermon, ``Scaling riemannian diffusion models,'' \emph{Advances in Neural Information Processing Systems}, vol.~36, pp. 80\,291--80\,305, 2023.

\bibitem{braun2024riemannian}
M.~Braun, N.~Jaquier, L.~Rozo, and T.~Asfour, ``Riemannian flow matching policy for robot motion learning,'' in \emph{2024 IEEE/RSJ International Conference on Intelligent Robots and Systems (IROS)}.\hskip 1em plus 0.5em minus 0.4em\relax IEEE, 2024, pp. 5144--5151.

\bibitem{spell2023mixture}
G.~P. Spell, S.~Ren, L.~M. Collins, and J.~M. Malof, ``Mixture manifold networks: a computationally efficient baseline for inverse modeling,'' in \emph{Proceedings of the AAAI Conference on Artificial Intelligence}, vol.~37, no.~8, 2023, pp. 9874--9881.

\bibitem{calder2022boundary}
J.~Calder, S.~Park, and D.~Slep{\v{c}}ev, ``Boundary estimation from point clouds: Algorithms, guarantees and applications,'' \emph{Journal of Scientific Computing}, vol.~92, no.~2, p.~56, 2022.

\bibitem{niyogi2011topological}
P.~Niyogi, S.~Smale, and S.~Weinberger, ``A topological view of unsupervised learning from noisy data,'' \emph{SIAM Journal on Computing}, vol.~40, no.~3, pp. 646--663, 2011.

\bibitem{bradley2000k}
P.~S. Bradley and O.~L. Mangasarian, ``K-plane clustering,'' \emph{Journal of Global optimization}, vol.~16, no.~1, pp. 23--32, 2000.

\bibitem{liu2023linear}
L.~Liu, J.~He, and Y.-H. Tsai, ``Linear regression on manifold structured data: the impact of extrinsic geometry on solutions,'' in \emph{Topological, Algebraic and Geometric Learning Workshops 2023}.\hskip 1em plus 0.5em minus 0.4em\relax PMLR, 2023, pp. 557--576.

\bibitem{liu2020low}
X.~Liu, H.~Huang, W.~Tang, T.~Zhang, and B.~Yang, ``Low-rank sparse tensor approximations for large high-resolution videos,'' in \emph{2020 19th IEEE International Conference on Machine Learning and Applications (ICMLA)}.\hskip 1em plus 0.5em minus 0.4em\relax IEEE, 2020, pp. 65--70.

\end{thebibliography}
\end{document}